\documentclass[sigconf]{acmart}

\usepackage{etoolbox}
\usepackage{epsfig, subfigure, amsmath}
\usepackage{algorithm}
\usepackage{algpseudocode}
\usepackage{balance}
\usepackage{color}
\usepackage{url}
\usepackage{booktabs} 

\apptocmd{\thebibliography}{}{}{}

\newcommand{\ie}{{\it i.e.}, }
\newcommand{\etal}{{\it et al.} }

\setcopyright{rightsretained} %

\acmConference[Mid-Southeast'18]{ACM Mid-Southeast conference}{November 2018}{Gatlinburg, Tennessee, USA}
\acmYear{2018}
\copyrightyear{2018}

\begin{document}

\title{Smart Weather Forecasting Using Machine Learning:\\
A Case Study in Tennessee}

\author{A H M Jakaria}
\affiliation{%
  \institution{Tennessee Tech University}
  \streetaddress{P.O. Box 5101}
  \city{Cookeville}
  \state{Tennessee}
  \postcode{38505}
}
\email{ajakaria42@students.tntech.edu}

\author{Md Mosharaf Hossain}
\affiliation{%
  \institution{Tennessee Tech University}
  \streetaddress{P.O. Box 5101}
  \city{Cookeville}
  \state{Tennessee}
  \postcode{38505}
}
\email{mhossain44@students.tntech.edu}

\author{Mohammad Ashiqur Rahman}
\affiliation{%
  \institution{Tennessee Tech University}
  \streetaddress{P.O. Box 5101}
  \city{Cookeville}
  \state{Tennessee}
  \postcode{38505}
}
\email{marahman@tntech.edu}

\renewcommand{\shortauthors}{A. Jakaria et al.}

\begin{abstract}
Traditionally, weather predictions are performed with the help of large complex models of physics, which utilize different atmospheric conditions over a long period of time. These conditions are often unstable because of perturbations of the weather system, causing the models to provide inaccurate forecasts. The models are generally run on hundreds of nodes in a large High Performance Computing (HPC) environment which consumes a large amount of energy. In this paper, we present a weather prediction technique that utilizes historical data from multiple weather stations to train simple machine learning models, which can provide usable forecasts about certain weather conditions for the near future within a very short period of time. The models can be run on much less resource intensive environments. The evaluation results show that the accuracy of the models is good enough to be used alongside the current state-of-the-art techniques. Furthermore, we show that it is beneficial to leverage the weather station data from multiple neighboring areas over the data of only the area for which weather forecasting is being performed.
\end{abstract}


\maketitle

\section{Introduction}
\label{Sec:Introduction}

Weather conditions around the world change rapidly and continuously. Correct forecasts are essential in today's daily life. From agriculture to industry, from traveling to daily commuting, we are dependent on weather forecasts heavily. As the entire world is suffering from the continuous climate change and its side effects, it is very important to predict the weather without any error to ensure easy and seamless mobility, as well as safe day to day operations.

The current weather prediction models heavily depend on complex physical models and need to be run on large computer systems involving hundreds of HPC nodes. The computational power of these large systems is required to solve the models that describe the atmosphere. Despite using these costly and complex devices, there are often inaccurate forecasts because of incorrect initial measurements of the conditions or an incomplete understanding of atmospheric processes. Moreover, it generally takes a long time to solve complex models like these.
%

As weather systems can travel a long way over time in all directions, the weather of one place depends on that of others considerably~\cite{amazon}. In this work, we propose a method to utilize surrounding city's historical weather data along with a particular city's data to predict its weather condition. We combine these data and use it to train simple machine learning models, which in turn, can predict correct weather conditions for the next  few days. These simple models can be run on low cost and less resource-intensive computing systems, yet can provide quick and accurate enough forecasts to be used in our day-to-day life. In this work, we present a case study on the city of Nashville in Tennessee, USA, which is known for its fluctuating weather patterns, and show that our simple model can provide reliable weather forecasts for this city.

The major contributions of this paper include: 
\begin{enumerate}
\item The utilization of machine learning in prediction of weather conditions in short periods of time, which can run on less resource-intensive machines.
\item Implementation of automated systems to collect historical data from a dedicated weather service.
\item Thorough evaluation of the proposed technique and comparison of several machine learning models in the prediction of future weather conditions.
\end{enumerate}

The rest of this  paper is organized as follows: Section~\ref{Sec:Background} gives an overview of machine learning in weather forecasting, as well as the related works. In Section~\ref{Sec:Technique}, we present the methodology of the proposed idea, which includes the methods to pull data from a weather station. We illustrate training and test data collection and their preprocessing in Section~\ref{Sec:Dataset}. Section~\ref{Sec:Results} shows the evaluation results of several machine learning techniques. We conclude the paper in Section~\ref{Sec:Conclusion}.

\section{Background}
\label{Sec:Background}

This section briefly presents how machine learning can be used in weather forecasting and the related works in the literature on this fast growing research topic.

\subsection{Machine Learning for Weather Forecasting} 
Machine learning is a data science technique which creates a model from a training dataset. A model is basically a formula which outputs a target value based on individual weights and values for each training variable. In each record, corresponding weights (sometimes between 0 and 1) to each variable tells the model how that variable is related to the target value. There must be sufficient amount of training data to determine the best possible weights of all the variables. When the weights are learned as accurately as possible, a model can predict the correct output or the target value given a test data record. 

Utilizing simple machine learning techniques allow us be relieved from the complex and resource-hungry weather models of traditional weather stations. It has immense possibilities in the realm of weather forecasting~\cite{next}. Such a forecasting model can be offered to the public as web services very easily~\cite{azure}.

\subsection{Related Works} 
Machine learning in weather forecasting is a recent trend in the literature. There are several works which discuss this topic.

Holmstrom \etal proposed a technique to forecast the maximum and minimum temperature of the next seven days, given the data of past two days~\cite{holmstrommachine}. They utilized a linear regression model, as well as a variation of a functional linear regression model. They showed that both the models were outperformed by professional weather forecasting services for the prediction of up to seven days. However, their model performs better in forecasting later days or longer time scales. A hybrid model that used neural networks to model the physics behind weather forecasting was proposed by Krasnopolsky and Rabinivitz~\cite{krasnopolsky2006complex}. Support vector machines was utilized for weather prediction as a classification problem by Radhika \etal~\cite{radhika2009atmospheric}. A data mining based predictive model to identify the fluctuating patterns of weather conditions was proposed in~\cite{yadav2016weather}. The patterns from historical data is used to approximate the upcoming weather conditions. The proposed data model uses Hidden Markov Model for prediction and $k$-means clustering for extracting weather condition observations. Grover \etal studied weather prediction via a hybrid approach, which combines discriminatively trained predictive models with deep neural networks that models the joint statistics of a set of weather-related variables~\cite{grover2015deep}.

Montori \etal used the concept of crowdsensing, where participating users share their smart phone data to environmental phenomenons~\cite{montori2017collaborative}. They introduced an architecture named SenSquare, which handles data from IoT sources and crowdsensing platforms, and display the data unifiedly to subscribers. This data is used in smart city environment monitoring. However, none of these works use the idea of combining data from neighboring places. 

\begin{figure}
	\centering
	\includegraphics[keepaspectratio,scale=0.42]{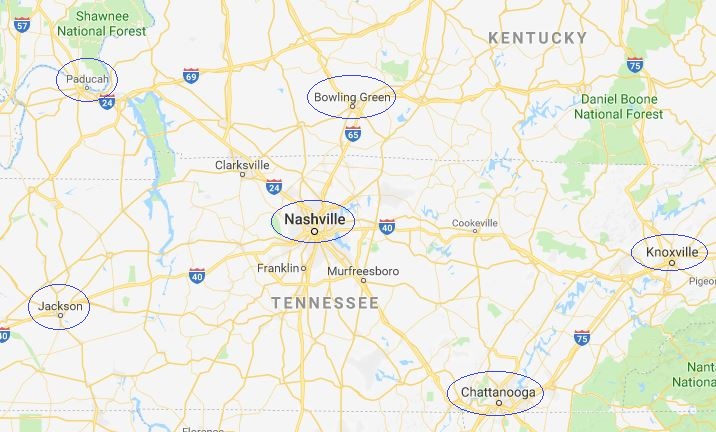}
	\caption{Google map showing the city of Nashville and its surrounding cities. The circled cities can have impact on the weather systems of Nashville.}
    \vspace{-12pt}
	\label{Fig_Map}
\end{figure}

\section{Technique Overview}
\label{Sec:Technique}

In this section, we present an overview of our proposed technique.

\subsection{Methodology}
In this case study, we aim to use ML techniques to predict the temperature of the next day at any particular hour of the city of Nashville, Tennessee, based on the weather data of the current day of this city and a couple of its surrounding cities. 

First, we combine the weather observations at a particular timestamp from all the cities that we consider to create a single record. That is, each record in the data will contain the temperature, humidity, wind direction, atmospheric pressure, condition, etc. data of all the cities. The target variable for this record is assigned as the temperature at the same timestamp of the next day. Thus, we predict the temperature of the next day given the weather observation of the current day. 

Fortunately, by the same way, we can also predict any other weather variable such as humidity, rainfall, wind speed and direction, visibility, etc. of the next day, as well as the next few days. However, we restrict our study for predicting the temperature only in this research.

\subsection{Machine Learning Techniques}
In this research, as the predicted outcomes are continuous numeric values, temperature in our case, we use regression technique. We find that Random Forest Regression (RFR) is the superior regressor, as it ensembles multiple decision trees while making decision. In addition, we show comparison of several other state-of-the-art ML techniques with the RFR technique. The incorporated regression techniques are Ridge Regression (Ridge), Support Vector (SVR), Multi-layer Perceptron (MLPR), and Extra-Tree Regression (ETR).


\section{Dataset}
\label{Sec:Dataset}
Once we collect the data, we split the raw data in training and test set. However, the target variable is always the next day hourly temperature for Nashville. The Training set contains two months of weather data starting from the 1st day of July, 2018. In contrast, the test set contains 7 days of data starting from September 1, 2018 and ending on September 7, 2018. Essentially, the trained model predicts hourly temperature of the 2nd September while inputting 1st September as test data. Similarly, temperature of September 3rd will be predicted based on data from September 2nd, and so on.
\subsection{Dataset from Weather Station}
\begin{figure*}[t]
    \begin{center}   	
       \subfigure[]{
            \label{1adding_cities}
            \includegraphics[scale=0.375, keepaspectratio=true]{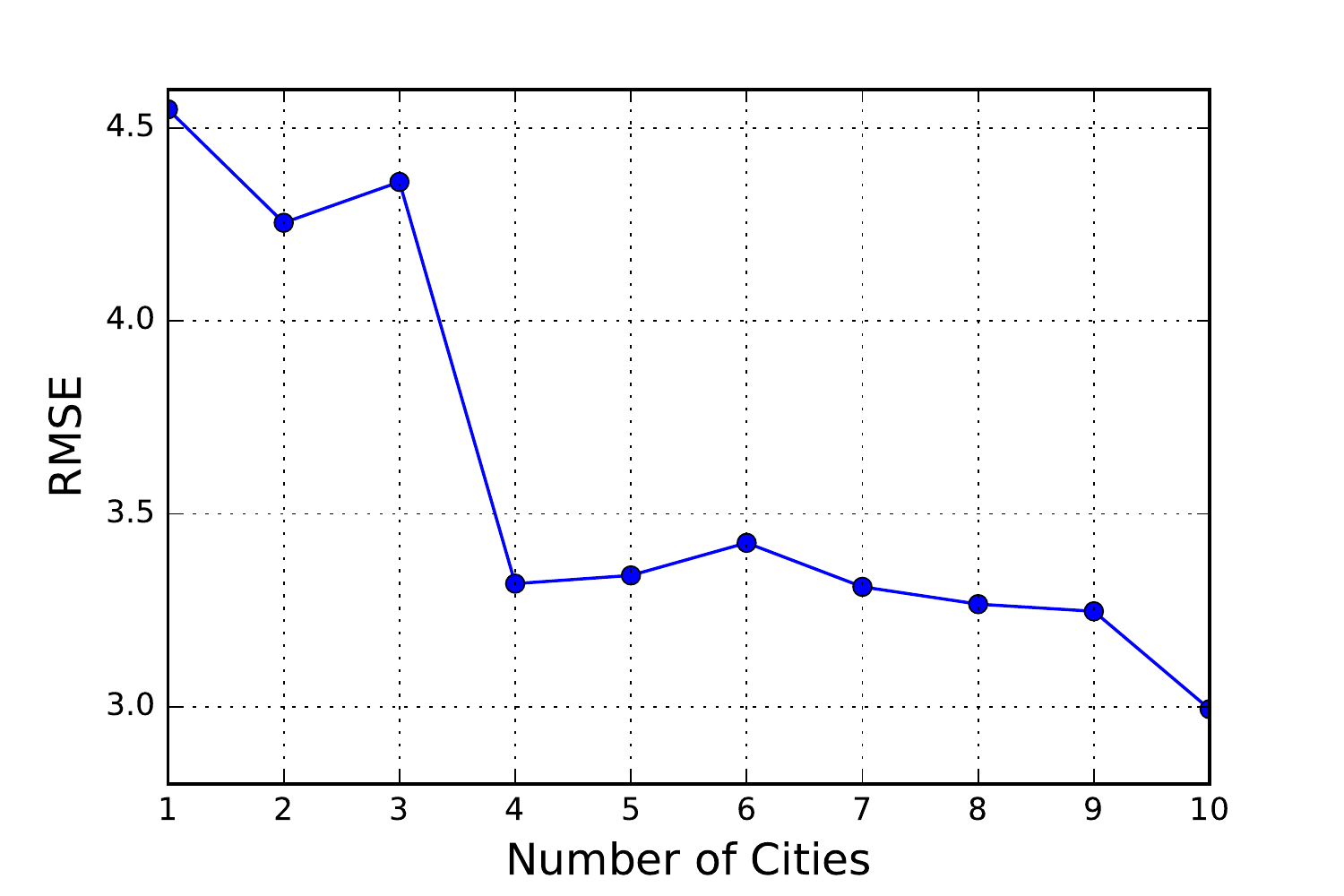}
       }
       \subfigure[]{
            \label{2weeks}
            \includegraphics[scale=0.375, keepaspectratio=true]{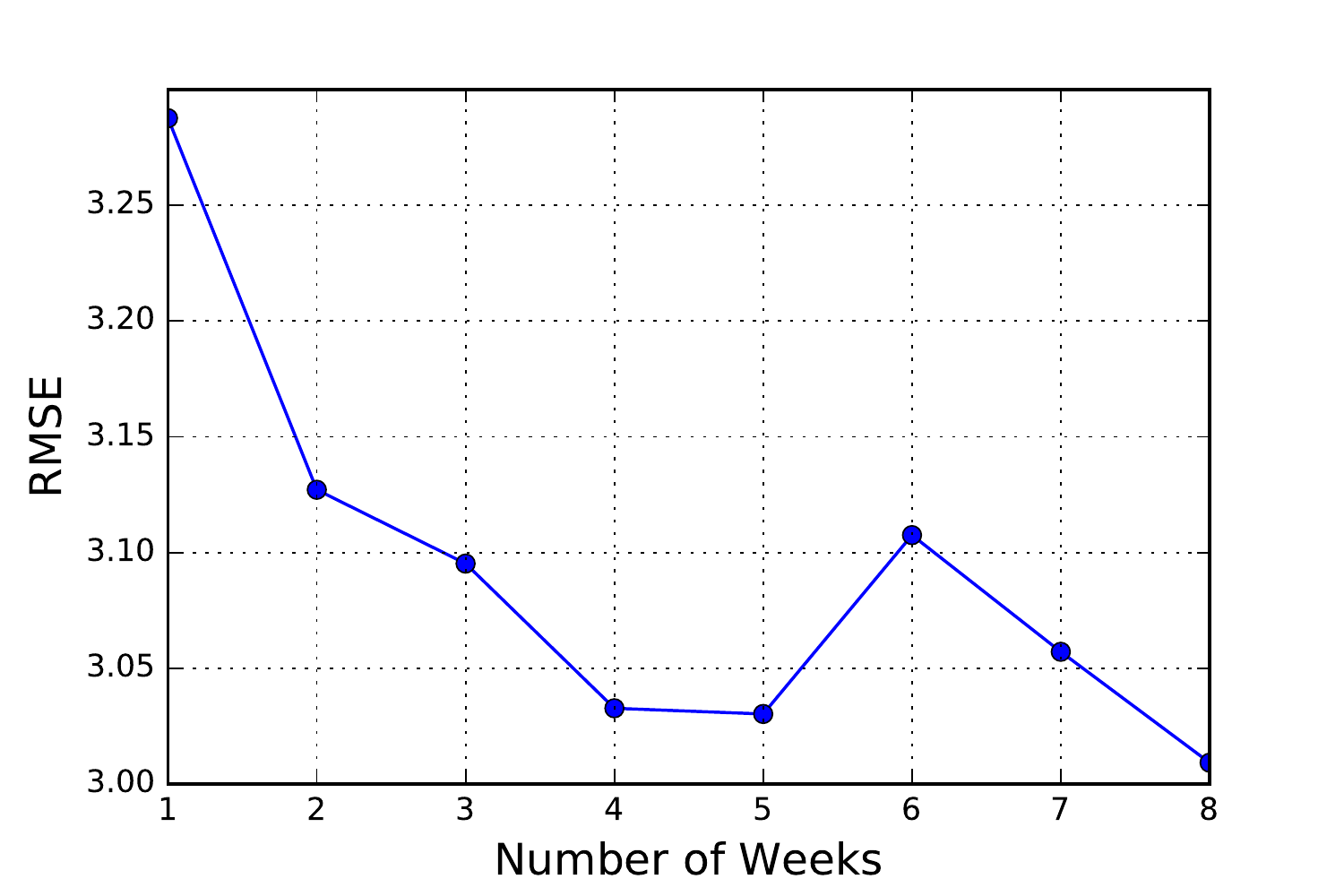}
       }
       \subfigure[]{
            \label{3ml_techniques_bar}
            \includegraphics[scale=0.375, keepaspectratio=true]{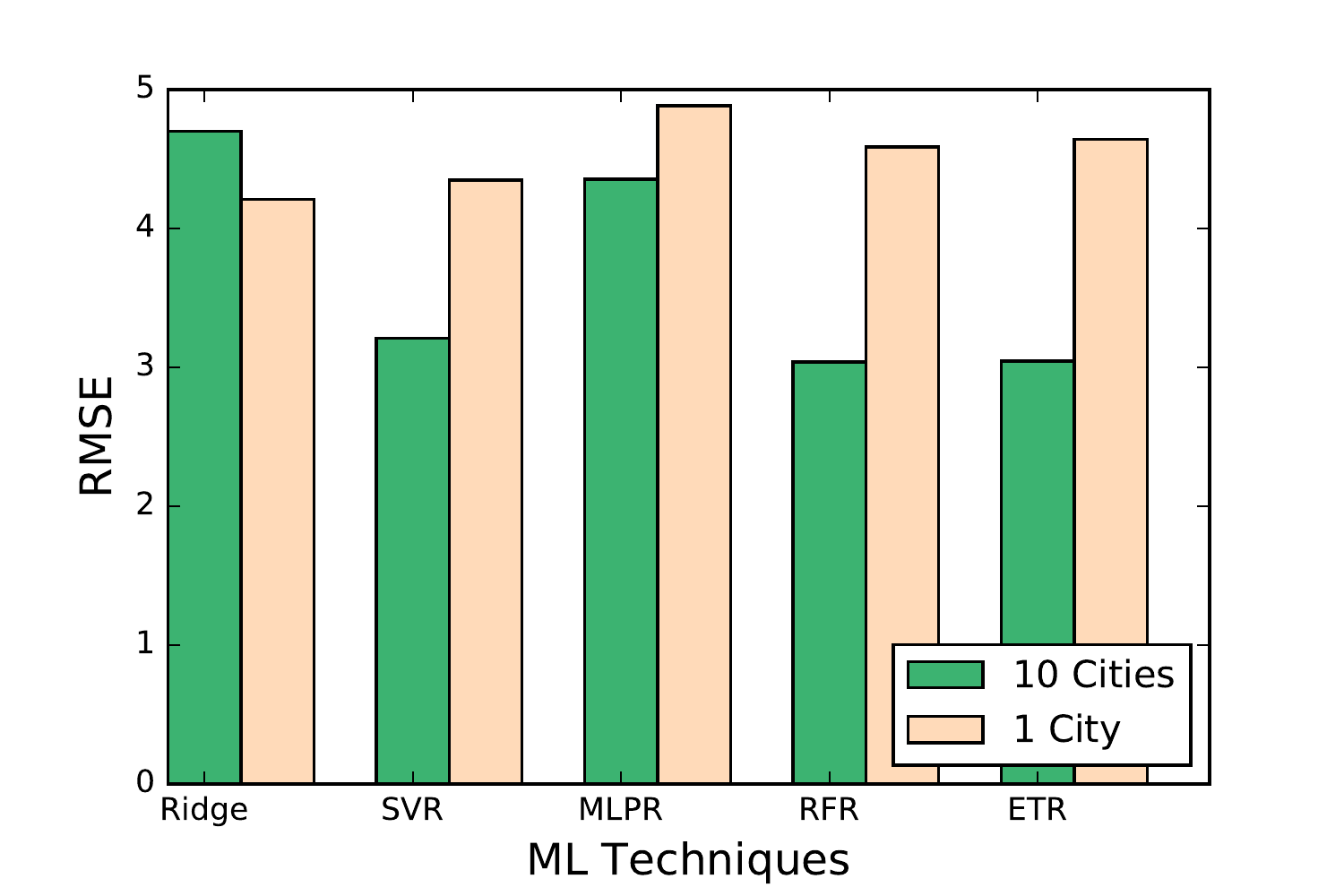}
       }
    \end{center}
    \vspace{-12pt}
    \caption{(a) RMSE on test set while considering neighboring cities, (b) RMSE on test set with increasing training size, (c) RMSE on test set for different ML models}
    \vspace{-12pt}
    \label{rmse}
\end{figure*}

We collected real weather data for the city of Nashville from {\em wunderground.com}~\cite{wunderground}, as well as nine more cities around Nashville: Knoxville, Chattanooga, Jackson, Bowling Green, Paducah, Birmingham, Atlanta, Florence, and Tupelo (See Fig.~\ref{Fig_Map}). For a given place and date, the wunderground API returns a list of weather observations data~\cite{wunderground}. A historical JSON query result consists of different weather parameters such as temperature, humidity, dew point, atmospheric pressure, wind direction, etc. at each hour of a day for a particular city. We collect this for the required number of days for different experiments.

We developed a program to convert the JSON data to a text file, where each record (row) corresponds to a particular timestamp. We skip any data from the JSON results that do not have data for all the ten cities for a particular timestamp.

\subsection{Data Preprocessing}
After having the raw data from `wunderground', we make sure that each row (record) in the dataset has records for all ten cities for a particular timestamp. We eliminate any feature with empty or invalid data while creating the dataset. Also, we convert the categorical features in the dataset, such as wind direction and condition, into dummy/indicator variables using a technique called `One Hot Encoding'~\cite{onehot}. We perform this conversion prior to the separation of training and test data. This is because, in both training and test data, we need the same number of feature variables. If we do this conversion after the separation, then there remains no guarantee that both of them will have all the categorical values for the corresponding features. If the number of categorical values for training and test sets is not the same, then the conversion yields to different number of features for these sets. That is why we need to perform this conversion before separation of training and test datasets.

Furthermore, we perform mean scaling $x \leftarrow \frac{x - \mu}{\sigma}$ to all the continuous variables so that the variables possess approximately zero mean, which in practice, reduces computational cost while training the models.

\section{Results}
\label{Sec:Results}

In this section, we present a thorough evaluation of our models trained with the weather station data. First set of results show prediction accuracy while increasing training data by adding more neighboring cities, and by adding more weeks. Second set of results mostly emphasize the noticeable performance improvement of our models when neighboring cities are included in the training data.

\subsection{Performance Measure}
In all our experiments, we use the root mean squared error (RMSE) to evaluate our models. The calculation of RMSE is pretty straight forward shown in Equation \ref{eq_rmse}.
\begin{equation}
\label{eq_rmse}
\mathit{RMSE} = \sqrt{\frac{\sum_{t=1}^{n}~(\hat{y}_t - y_t)^2}{n}}
\end{equation}
Where $n$ is the number of test examples. $\hat{y}_t$ and $y_t$ are predicted temperature and actual temperature, respectively.

\begin{figure*}[t]
    \begin{center}   	
       \subfigure[]{
            \label{err}
            \includegraphics[scale=0.375, keepaspectratio=true]{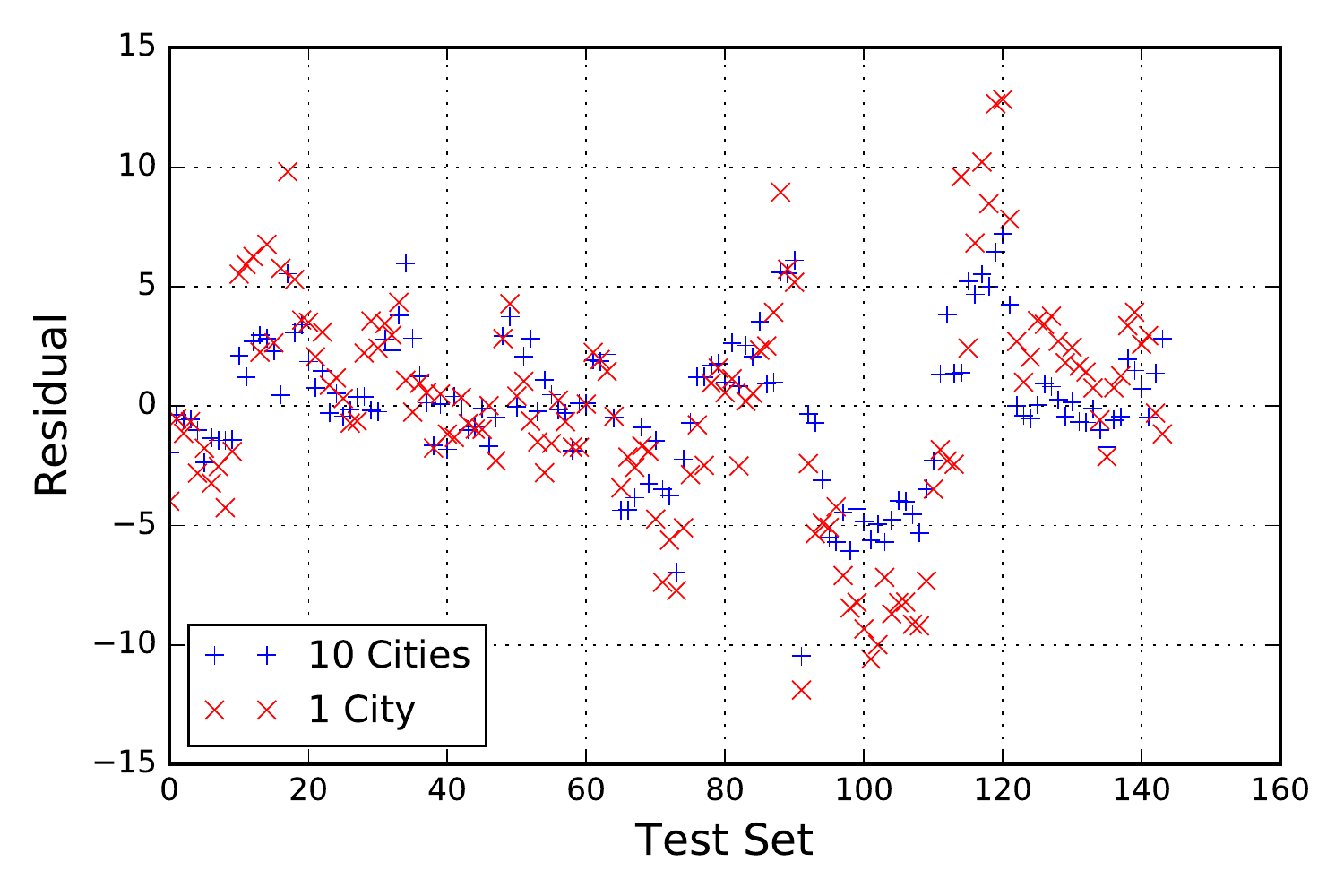}
       }
       \subfigure[]{
            \label{dist}
            \includegraphics[scale=0.375, keepaspectratio=true]{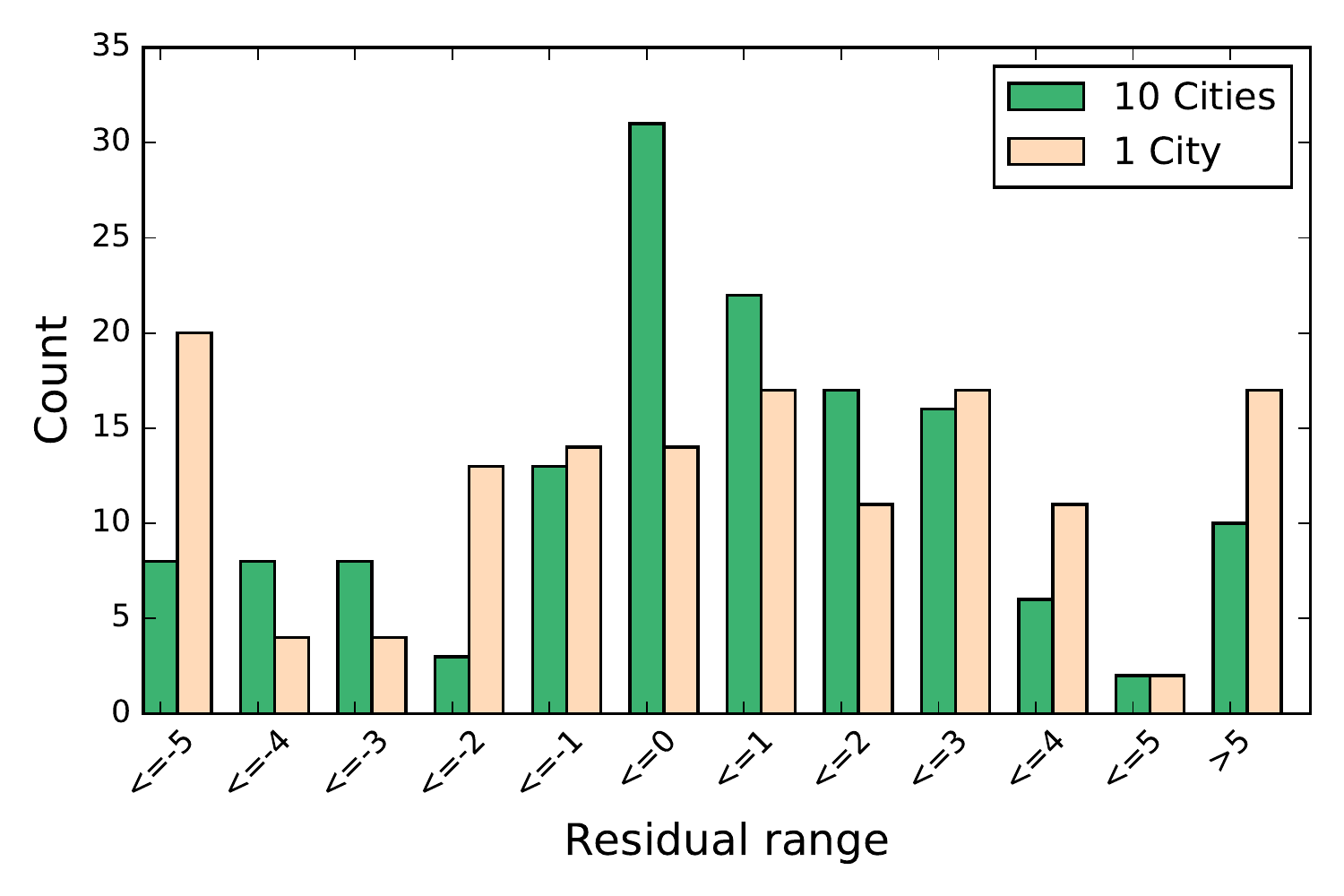}
       }
       \subfigure[]{
            \label{test-rmse}
            \includegraphics[scale=0.375, keepaspectratio=true]{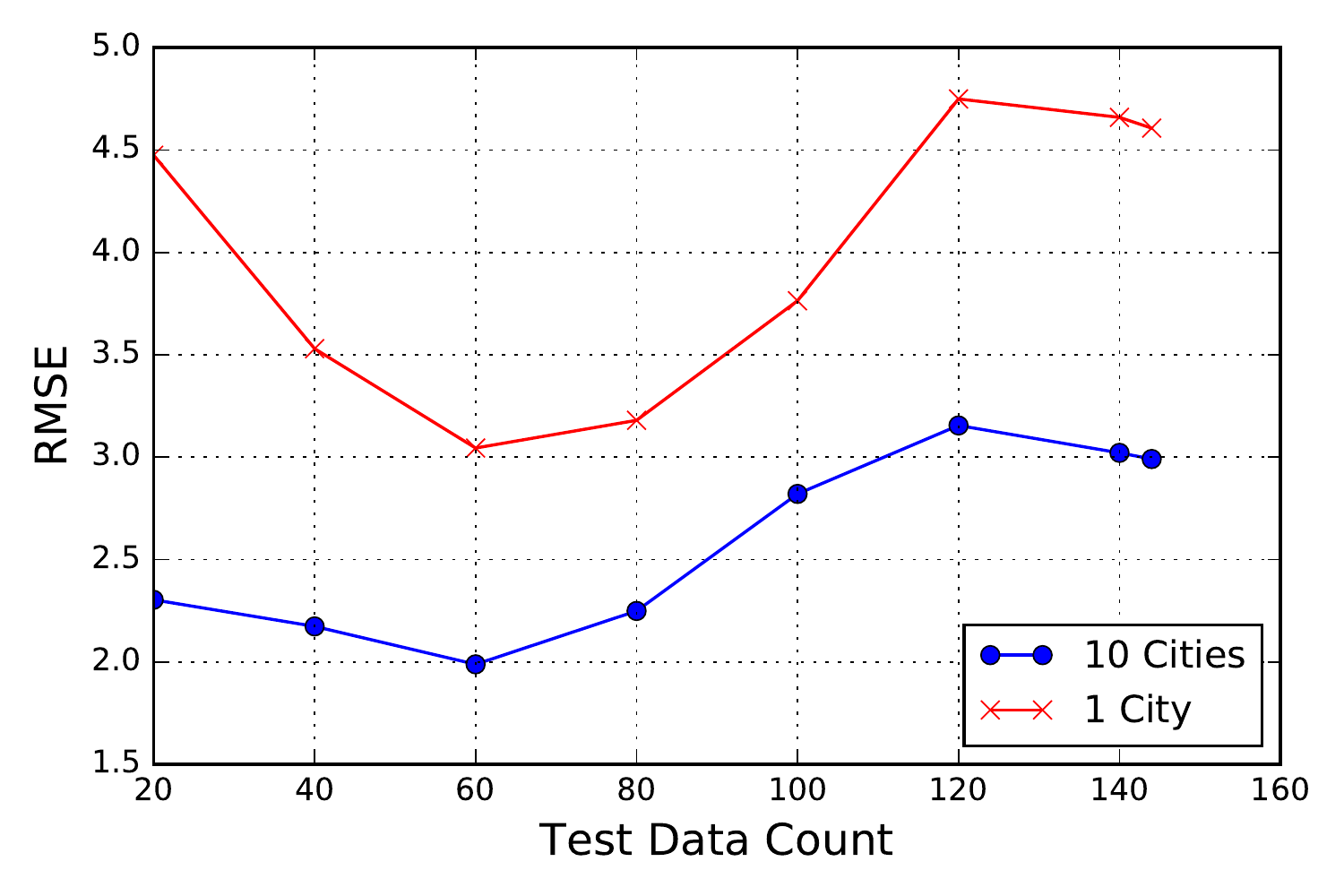}
       }
    \end{center}
    \vspace{-12pt}
    \caption{(a) Error in prediction of the test set, (b) distribution of errors in prediction of the test set, (c) RMSE for different number of test data.}
    \vspace{-12pt}
    \label{Fig_Err}
\end{figure*}

\subsection{Performance Comparison of Models}
At the very beginning of our experiment, we incrementally add more cities to observe the RMSE trend on the test data. Fig.~\ref{1adding_cities} exposes our expected trend, where we add nine other cities to train the RFR model. Once the model is trained with data of Nashville only, the observed RMSE is noticeably high, which is approximately $4.5$. As we add one more neighboring city, the model starts to perform well. However, when 3 neighboring cities are considered, model accuracy degrades slightly. The potential reason is that Nashville weather might not depend on all the neighboring cities essentially all the time. Some seasonal change or bad weather situation of a neighboring region might not affect a particular region throughout the whole year. However, once we add more cities, RMSE apparently continues to decrease. At the ten-city case, significantly low RMSE is observed, which is nearly $35\%$ less as compared to only one city data.

In Fig.~\ref{2weeks}, we show RMSE on test data for growing number of training data of all the ten cities. At first, only one week data is used as training set, that week is the immediate previous week of the test week. As we said earlier, the test data comprises of seven days of data starting from September 1, 2018. Thus, first week of training data represents the weather data starting from the 25th August and ending on the 31st August, 2018. Two weeks of training refers to the data of the previous two weeks of the test week, and so on. Using only one week, RMSE is noticeably high, nearly $3.3$. As we increase the number of weeks, RMSE drops considerably. At $5$-week situation, RMSE is nearly $3.03$. After that, we find a quick increase of RMSE as we add some more weeks. The possible reason is that in some weeks, weather condition might change abruptly, which influences the training model when we incorporate those data. However, when training set comprises of $8$ weeks of data, we find lowest RMSE on the test set.

Later on, we show comparison of different ML models on the test data in Fig.~\ref{3ml_techniques_bar}. Each model shows RMSE when training with ten cities, as well as when training with only one city (Nashville). The first ML technique we use is Ridge regression. In this particular regression model, both ten and single city situation show relatively high RMSE above $4.0$. After that, we use Support Vector Regressor (SVR). This technique shows pretty good accuracy for ten-city case compared to one city. The RMSE difference is near to $1$ for the two cases. Next, we use Multi-Layer Perceptron Regressor (MLPR), which is a two layer Neural Network, comprising with $100$ units in first hidden layer and $50$ units in second hidden layer. Unfortunately, the MLPR technique shows high RMSE in both ten and one city cases. Final two models are Random Forest Regressor (RFR) and Extra-Tree Regressor (ETR). Fortunately, in these two models, the ten-city case presents nearly similar RMSE close to $3.0$, which is the lowest among all the previous three models. On the contrary, in one city situation, both of these models show quite high RMSE. This explains the necessity of considering weather data of neighboring cities when predicting temperature of a particular city or region.

In Fig.~\ref{err}, we show the prediction errors or residuals, \ie the difference between the predicted target value and the actual target value, in the test set. As before, We plot two sets of data points for the same test set - one with training set from only the city of Nashville and the other one is including nine more surrounding cities. It can be observed that the data points for only one city are more dispersed than the data points for ten cities. This apparently shows that the possibility of error is much lower for every test point when we consider data from multiple cities.

The distribution of the residual is presented in Fig.~\ref{dist}, where we present 12 buckets of residual. The central bucket yields count of the test points where their residuals lie in the (-1, 0] range. The result reveals that it is better to use training data from multiple cities for training. The distribution of the residuals is clustered in central bucket for multiple cities, while the distribution of one city is more likely to end up in residual buckets far from central.

Fig.~\ref{test-rmse} presents the RMSE for test data of increasing size in each run. We start with $20$ test examples, where ten-city case shows $50\%$ less RMSE than the single-city case. Once the number of test examples reaches $60$, the ten-city case shows surprisingly less error. With bigger test sets beyond $60$, performance of both cases degrade, because weather data in some test data might be very dissimilar to that of training set. However, for bigger test size, It is apparent that the model performs better when it is trained with data from multiple surrounding cities compared to training with one city.


\section{Conclusion and Future Directions}
\label{Sec:Conclusion}

In this paper, we presented a technology to utilize machine learning techniques to provide weather forecasts. Machine learning technology can provide intelligent models, which are much simpler than traditional physical models. They are less resource-hungry and can easily be run on almost any computer including mobile devices. Our evaluation results show that these machine learning models can predict weather features accurately enough to compete with traditional models. We also utilize the historical data from surrounding areas to predict weather of a particular area. We show that it is more effective than considering only the area for which weather forecasting is done.

In future, we have plans to utilize low-cost Internet of Things (IoT) devices, such as temperature and humidity sensors, in collecting weather data from different parts of a city. The use of different sensors could increase the number of local features in the training dataset. This data, along with the weather station data, will further improve the performance of our prediction models.

\bibliographystyle{ACM-Reference-Format}
\bibliography{Weather}

\end{document}